\documentclass[review]{elsarticle}
\usepackage{graphicx}
\usepackage{hyperref}
\usepackage{adjustbox}
\usepackage{subfigure}
\usepackage{multirow}
\usepackage{xcolor}

\journal{Neurocomputing}

\usepackage{lipsum} 
\usepackage{fancyhdr}
\begin{document}
\newcommand{\etal}{\textit{et al}. }
\begin{frontmatter}

\title{Impact of Fully Connected Layers on Performance of Convolutional Neural Networks for Image Classification}

\author{S H Shabbeer Basha, Shiv Ram Dubey, Viswanath Pulabaigari, Snehasis Mukherjee}
\address{Indian Institute of Information Technology Sri City, Andhra Pradesh- 517646, India.}




\begin{abstract}
The Convolutional Neural Networks (CNNs), in domains like computer vision, mostly reduced the need for handcrafted features due to its ability to learn the problem-specific features from the raw input data.
However, the selection of dataset-specific CNN architecture, which mostly performed by either experience or expertise is a time-consuming and error-prone process. To automate the process of learning a CNN architecture, this paper attempts at finding the relationship between Fully Connected (FC) layers with some of the characteristics of the datasets. The CNN architectures, and recently datasets also, are categorized as deep, shallow, wide, etc. This paper tries to formalize these terms along with answering the following questions. (i) \textit{What is the impact of deeper/shallow architectures on the performance of the CNN w.r.t. FC layers?}, (ii) \textit{How the deeper/wider datasets influence the performance of CNN w.r.t. FC layers?}, and (iii) \textit{Which kind of architecture (deeper/shallower) is better suitable for which kind of (deeper/wider) datasets}. To address these findings, we have performed experiments with three CNN architectures having different depths. The experiments are conducted by varying the number of FC layers. We used four widely used datasets including CIFAR-10, CIFAR-100, Tiny ImageNet, and CRCHistoPhenotypes to justify our findings in the context of image classification problem. The source code of this work is available at \url{https://github.com/shabbeersh/Impact-of-FC-layers}.
\end{abstract}

\begin{keyword}
Convolutional Neural Networks \sep Fully Connected Layers \sep Image Classification \sep Shallow vs Deep CNNs \sep Wider vs Deeper Datasets.
\end{keyword}

\end{frontmatter}


\section{Introduction and Related Works}
\label{sec1}
The popularity of Convolutional Neural Networks (CNN) is growing significantly for various application domains related to computer vision, which include object detection \cite{lecun2015deep}, segmentation \cite{he2017mask}, localization \cite{hariharan2017object}, and many more in recent years. Despite the success of deep learning models, our theoretical understanding about neural networks remains limited. Careful selection of network width (number of neurons in FC layers, number of filters in convolution layers) and network depth (number of trainable layers) plays a vital role in designing deep neural networks in order to obtain better performance. 
In this paper, we made an attempt to find some of the factors which affect the performance of the CNN w.r.t. Fully Connected (FC) layers in the context of image classification. We have also studied the possible interrelationship between the presence of FC layers in CNN, the depth of the CNN, and the depth of the dataset.

Deep neural networks usually provide better results in the field of machine learning and computer vision compared to the handcrafted feature descriptors \cite{lecun2015deep}. From the available literature, it is apparent that every CNN architecture have one or more FC layers depending on the architecture's depth. To mention a few, AlexNet \cite{krizhevsky2012imagenet} consists of $5$ convolutional ($Conv$) layers and $3$ FC layers. The FC layers are placed after all the Conv layers. Zeiler and Fergus \cite{zeiler2014visualizing} made minimal changes to AlexNet with better hyper-parameter settings in order to generalize it over other datasets. This model is called ZFNet which also has $3$ FC layers along with $5$ convolution layers. In $2014$, Simonyan \etal\cite{simonyan2014very} further extended the AlexNet model to VGG-16 with $16$ learnable layers including $3$ FC layers towards the end of the architecture. Later on, many CNN models have been introduced with an increasing number of learnable layers. Szegedy \etal \cite{szegedy2015going} proposed a $22$-layer architecture called GoogLeNet, which has a single FC (output) layer. In $2015$, He \etal \cite{he2016deep} introduced ResNet with $152$ trainable layers where the last layer is fully connected. However, all the above CNN architectures are proposed for large-scale ImageNet dataset \cite{deng2009imagenet}. Recently, Basha \etal \cite{basha2018rccnet} proposed a CNN based classifier called RCCNet, which is responsible for classifying the routine colon cancer cells of dimension $32\times32\times3$. This CNN model has $7$ learnable layers including $3$ FC layers. 


\begin{figure}[!t]
\centering
\includegraphics[width=1 \textwidth]{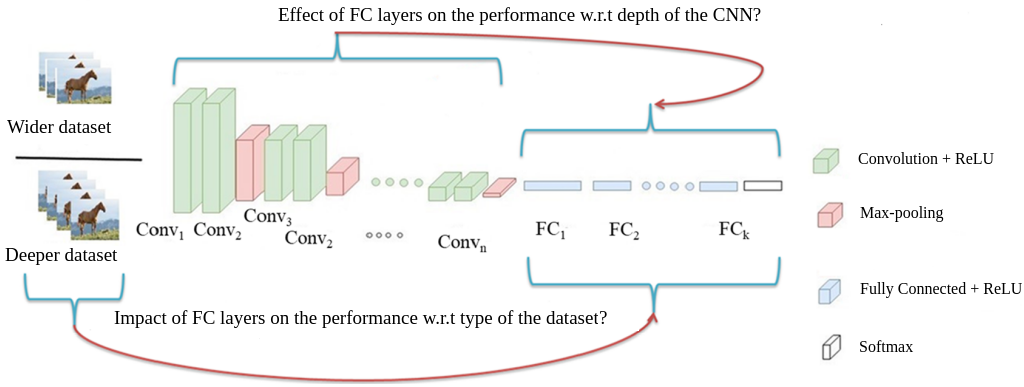}
\caption{The illustration of the effect of deeper/wider datasets and depth of CNN (i.e., the number of the Convolutional layers, $n$) over the number of FC layers (i.e., $k$). A typical plain CNN architecture has Convolutional (learnable), Max-pooling (non-learnable) and FC (learnable) layers. 
}
\label{a_motivational_fig}
\end{figure}


\textbf{Necessity of Fully Connected Layers in CNN:}
In a shallow CNN model, the features generated by the final convolutional layer correspond to a portion of the input image as its receptive field does not cover the entire spatial dimension of the image. Thus, few FC layers are mandatory in such a scenario. Despite their pervasiveness, the hyperparameters like the number of FC layers and number of neurons required in FC layers for a given CNN architecture to obtain better performance are not explored.

In a typical deep neural network, the FC layers comprise most of the parameters of the network. AlexNet has $60$ million parameters, out of which $58$ million parameters correspond to the FC layers \cite{krizhevsky2012imagenet}. Similarly, \mbox{VGGNet} has a total of $138$ million parameters, out of which $123$ million parameters belong to FC layers \cite{simonyan2014very}. This huge number of trainable parameters in FC layers are required to fit complex nonlinear discriminant functions in the feature space into which the input data elements are mapped.
However, this large number of parameters may result in over-fitting the classifier (CNN). To reduce the amount of over-fitting, Xu \etal \cite{xu2018overfitting} proposed a CNN architecture called \mbox{SparseConnect} where the connections between {\em FC} layers are sparsed.


The effect of deep or shallow networks on different kind of datasets is well explored in the literature to study the behavioral interrelationship between depth of dataset and the CNN \cite{mhaskar2016deep, mhaskar2016learning}. Mhaskar \etal \cite{mhaskar2016deep} extended a framework for their previous work \cite{mhaskar2016learning} to investigate when deep networks are better than shallow networks using a Directed Acyclic Graph (DAG). Montufar \etal \cite{montufar2014number} performed a study to find the complexity of the functions computable by deep neural networks with linear activations. 



To the best of our knowledge, no effort has been made in the literature to analyze the role of FC layers in CNN for image classification. In this paper, we investigate the impact of FC layers on the performance of the CNN model with a rigorous analysis from various aspects. In brief, the contributions of this paper are summarized as follows.
\begin{itemize}
\item We perform a systematic study to observe the effect of deeper/shallower architectures on the performance of CNNs with varying number of FC layers.

\item We observe the effect of deeper/wider datasets on the performance of CNN w.r.t. FC layers.

\item We generalize one important finding of Bansal \etal \cite{bansal2017s} to choose deeper or shallow architecture based on the depth of the dataset. In \cite{bansal2017s}, they have reported the same in the context of face recognition, Whereas, we made a rigorous study to generalize this observation over different kinds of datasets.
\item To make the empirical justification of our findings, we have conducted the experiments on different modalities (i.e., natural and bio-medical images) of image datasets like CIFAR-10, CIFAR-100 \cite{krizhevsky2009learning}, Tiny ImageNet \cite{tinyimagenet}, and CRCHistoPhenotypes  \cite{sirinukunwattana2016locality}.
\end{itemize}

Next, we illustrate the developed deep and shallow CNN architectures to conduct the experiments in Section \ref{Developed_CNN_models}. Experimental setup including training details, evaluation criteria, and datasets are discussed in Section \ref{experimental_setup}. Section \ref{results_analysis} presents a detailed study of the observations found in this paper. At last, Section \ref{conclusion}  
concludes the paper.

\section{Developed CNN Architectures}
\label{Developed_CNN_models}
The main objective of this paper is to analyze the impact of different hyperparameters realted to FC layers (the number of FC layers and the number of neurons) over the performance. Inter-dependency between the characteristics of both the datasets and the networks are explored w.r.t. FC layers as shown in Fig. \ref{a_motivational_fig}. In order to conduct a rigorous experimental study, we have implemented four CNN models among which three CNN models are plain architectures. Another model involves skip connections as in ResNet \cite{he2016deep}. These models are termed as CNN-1, CNN-2, and CNN-3.

\textbf{Deep and Shallow CNNs}: As per the published literature \cite{ba2014deep, montufar2014number}, a neural network is referred to as shallow if it has single fully connected (hidden) layer. Whereas, a deep CNN consists of convolution layers, pooling layers, and FC layers. However, in this paper, we assume a CNN model $N_1$ as deep/shallow compared to another CNN model $N_2$, if the number of trainable layers in $N_1$ is more/less than $N_2$, respectively. 


\subsection{CNN-1 Architecture}
AlexNet \cite{krizhevsky2012imagenet} is well-known CNN architecture, which won the first ImageNet Large Scale Visual Recognition Challenge (ILSVRC) in 2012 \cite{russakovsky2015imagenet} with a huge performance gain as compared to the best results of that time using handcrafted features. The AlexNet architecture was proposed for the images of dimension $227 \times 227 \times 3$, we made minimal changes to the model to fit for low-resolution images. We name this model as CNN-1. Initially, the input image dimension is up-sampled from $32\times32\times3$ to $35\times35\times3$ in the case of \mbox{CRCHistoPhenotypes} \cite{sirinukunwattana2016locality}, CIFAR-10, CIFAR-100 \cite{krizhevsky2009learning} datasets. Whereas, the images of Tiny ImageNet dataset \cite{tinyimagenet} are down-sampled from $64\times64\times3$ to $35\times35\times3$. The $1^{st}$ convolutional layer $Conv1$ produces $31\times31\times96$ dimensional feature vector by applying $96$ filters of dimension $5\times5\times3$. The $Conv1$ layer is followed by another Convolution layer ($Conv2$), which produces $27\times27\times256$ dimensional feature map by convolving $256$ filters of size $5\times5\times96$. The remaining layers of the CNN-1 model are similar to the AlexNet architecture proposed in \cite{krizhevsky2012imagenet}. The CNN-1 model with a single FC layer (i.e., the output FC layer) consists of following number of trainable parameters, $4,152,906$ for CIFAR-10 dataset, $8,046,756$ for CIFAR-100 dataset, $12,373,256$ for Tiny ImageNet dataset, and $3,893,316$ for CRCHistoPhenotypes dataset. Note that, the number of trainable parameters are different for each dataset due to the different number of classes present in the datasets which leads to the varying number of trainable parameters in the output FC layer. The detailed specifications of the CNN-1 model are given in Table \ref{table_CNN-1}.

\begin{table}[!t]
\caption{The CNN-1 architecture having $5$ $Conv$ layers. The $S$, $P$, and $BN$ denote stride, padding, and batch normalization, respectively. The output (FC) layer has $10$, $100$, $200$, and $4$ neurons in the case of CIFAR-10, CIFAR-100, \mbox{Tiny ImageNet}, and CRCHistoPhenotypes datasets, respectively.}
\label{table_CNN-1}
\begin{center}
\scalebox{1}{
\begin{tabular}{l l }
\hline Input: &Image dimension ($35\times35\times3$) \\
\hline
\hline
[layer $1$]& Conv. $(5,5,96)$, S=$1$, P=$0$; ReLU; BN;\\
\hline
[layer $2$]& Conv. $(5,5,256)$, S=$1$, P=$0$;  ReLU; BN;\\
\hline
[layer $3$]& Pool., S=$2$, P=$0$; \\
\hline
[layer $4$]& Conv. $(3,3,384)$, S=$1$, P=$1$; ReLU;\\
\hline
[layer $5$]& Conv. $(3,3,384)$, S=$1$, P=$1$; ReLU;\\
\hline
[layer $6$]& Conv. $(3,3,256)$, S=$1$, P=$1$; ReLU;\\
\hline
[layer $7$]& Flatten; $43264$;\\
\hline
\hline
Output:& (FC layer) Predicted Class Scores\\
\hline
\end{tabular}
}
\end{center}
\end{table}

\begin{table}[!t]
\caption{The CNN-2 model having 10-$Conv$ layers. The $S$, $P$, $BN$, and $DP_f$ denote the stride, padding, batch normalization, and dropout with a factor of $f$. The output layer has $10$, $100$, $200$, and $4$ neurons in the case of CIFAR-10 
, CIFAR-100 
, Tiny ImageNet, and CRCHistoPhenotypes 
datasets, respectively.}
\label{12_layer_model}
\begin{center}
\scalebox{1}{
\begin{tabular}{l l }
\hline Input: &Image dimension ($32\times32\times3$) \\
\hline
\hline
[layer $1$]& Conv. $(3,3,64)$, S=$1$, P=$1$;  ReLU; BN, DP$_{0.3}$\\
\hline
[layer $2$]& Conv. $(3,3,64)$, S=$1$, P=$1$;  ReLU; BN;\\
\hline
[layer $3$]& Pool., S=$2$, P=$0$; \\
\hline
\hline
[layer $4$]& Conv. $(3,3,128)$, S=$1$, P=$1$;  ReLU; BN, DP$_{0.4}$\\
\hline
[layer $5$]& Conv. $(3,3,128)$, S=$1$, P=$1$;  ReLU; BN;\\
\hline
[layer $6$]& Pool., S=$2$, P=$0$; \\
\hline
\hline
[layer $7$]& Conv. $(3,3,256)$, S=$1$, P=$1$;  ReLU; BN, DP$_{0.4}$\\
\hline
[layer $8$]& Conv. $(3,3,256)$, S=$1$, P=$1$;  ReLU; BN;\\
\hline
[layer $9$]& Pool., S=$2$, P=$0$; \\
\hline
\hline
[layer $10$]& Conv. $(3,3,512)$, S=$1$, P=$1$;  ReLU; BN, DP$_{0.4}$\\
\hline
[layer $11$]& Conv. $(3,3,512)$, S=$1$, P=$1$;  ReLU; BN;\\
\hline
[layer $12$]& Pool., S=$2$, P=$0$; \\
\hline
\hline
[layer $13$]& Conv. $(3,3,512)$, S=$1$, P=$1$;  ReLU; BN, DP$_{0.4}$\\
\hline
[layer $14$]& Conv. $(3,3,512)$, S=$1$, P=$1$;  ReLU; BN;\\
\hline
[layer $15$]& Pool., S=$2$, P=$0$; \\
\hline
\hline
[layer $16$]& Flatten; 512;\\
\hline
\hline
Output:& (FC layer) Predicted Class Scores\\
\hline
\end{tabular}
}
\end{center}
\end{table}

\begin{figure}
\centering
    \subfigure[]{\includegraphics[width=0.3\textwidth]{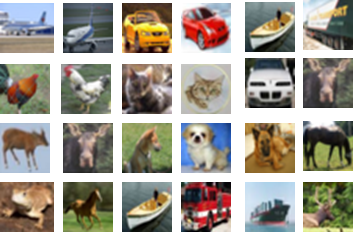}} \label{CIFAR-10_dataset}
    \subfigure[]{\includegraphics[width=0.3\textwidth]{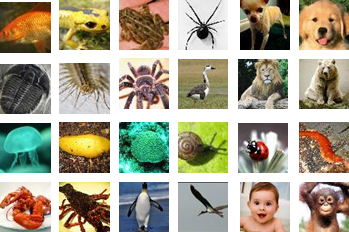}}\label{Tinyimagenet_dataset}
   \subfigure[]{\includegraphics[width=0.3\textwidth]{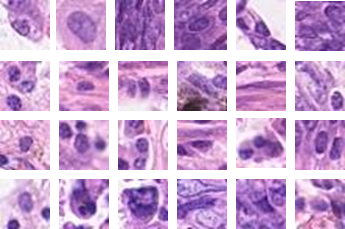}}\label{CRC_dataset}
 \caption{(a) A few sample images from CIFAR-10/100 dataset \cite{krizhevsky2009learning}. (b) A random sample images from Tiny ImageNet dataset \cite{tinyimagenet}. (c) Example images from CRCHistoPhenotypes dataset \cite{sirinukunwattana2016locality} with each row represents the images from one category.}
\label{image_samples}
\end{figure}

\subsection{CNN-2 Architecture} 
Another CNN model is designed based on the CIFAR-VGG \cite{liu2015very} model by removing some $Conv$ layers from the model. We name this model as CNN-2. The CNN-2 has $6$ blocks, where first $5$ blocks have two consecutive $Conv$ layers followed by a $Pool$ layer. Finally, the sixth block has a FC (output) layer which generates the class scores. The input to this model is an image of dimension $32\times32\times3$. To meet this requirement, images of the Tiny ImageNet dataset are down-sampled from $64\times64\times3$ to $32\times32\times3$. The CNN-2 architecture corresponds to $9416010$, $9462180$, $9513480$, and $9412932$ trainable parameters in the case of CIFAR-10, CIFAR-100, Tiny ImageNet, and CRCHistoPhenoTypes datasets, respectively. The CNN-2 model specifications are given in Table \ref{12_layer_model}.

\subsection{CNN-3 Architecture} 
Most of the popular CNN models like AlexNet \cite{krizhevsky2012imagenet}, VGG-16 \cite{simonyan2014very}, GoogLeNet \cite{szegedy2015going}, and many more were proposed for high dimensional image dataset called ImageNet \cite{deng2009imagenet}. On the other hand, the low dimensional image datasets such as CIFAR-10/100 have rarely got benefited from the CNNs. Liu \etal \cite{liu2015very} introduced CIFAR-VGG architecture, which is basically a $16$ layer deep CNN architecture proposed for CIFAR-10. We have utilized CIFAR-VGG model as the third deep neural network to observe the impact of FC layers in CNN and named as CNN-3 in this paper. The input to this model is an image of dimension $32\times32\times3$. To meet this requirement, images of the Tiny ImageNet dataset are down-sampled from $64\times64\times3$ to $32\times32\times3$. The CNN-3 architecture with a single FC (output) layer corresponds to $14728266$, $14774436$, $14825736$, and $14725188$ trainable parameters in the case of CIFAR-10 \cite{krizhevsky2009learning}, CIFAR-100 \cite{krizhevsky2009learning}, Tiny ImageNet \cite{tinyimagenet}, and CRCHistoPhenotypes \cite{sirinukunwattana2016locality} datasets, respectively.


\section{Experimental Setup}
\label{experimental_setup}
  This section describes the experimental setup including the training details, datasets used for the experiments, and the evaluation criteria to judge the performance of the CNN models.

\subsection{Training details}
The classification experiments are conducted on different modalities of image datasets to provide the empirical justifications of our findings in this paper. The initial value of the learning rate is $0.1$ and it is decreased by a factor of $2$ for every $20$ epochs. The Rectified Linear Unit ($ReLU$) based non-linearity \cite{krizhevsky2012imagenet} is used as the activation function after every $Conv$ and FC layer (except the output FC layer) in all the CNN models discussed in section \ref{Developed_CNN_models}. The Batch Normalization (i.e., $BN$) \cite{ioffe2015batch} is employed after $ReLU$ of each $Conv$ and FC layer, except final FC layer in CNN-2 and CNN-3 architectures. Whereas, in the case of CNN-1, the Batch Normalization is used only with the first two $Conv$ layers as mentioned in Table \ref{table_CNN-1}. To reduce the amount of over-fitting, we have used a popular regularization method called Dropout (i.e., $DP$) \cite{srivastava2014dropout} after some Batch-Normalization layers as summarized in Table \ref{12_layer_model} for CNN-2. For CNN-3, the $DP$ layers are used as per the CIFAR-VGG model \cite{liu2015very}. 
In order to find the impact of fully connected (FC) layers on the performance of CNN, any added FC layer has the $ReLU$, $BN$ and $DP$ by default.
Along with dropout, various data augmentations techniques like rotation, horizontal flip, and vertical flip are also applied to reduce the amount of over-fitting. The implemented CNN architectures are trained for $250$ epochs using Stochastic Gradient Descent (SGD) optimizer with a momentum of $0.9$.

\subsection{Evaluation criteria}
To evaluate the performance of the developed CNN models (i.e., CNN-1, CNN-2, and CNN-3), we have considered the classification accuracy as the performance evaluation metric. 

\subsection{Datasets}
\label{datasets}
To find out the empirical observations addressed in this paper, we have conducted the experiments on different modalities of datasets such as CIFAR-$10$ \cite{krizhevsky2009learning}, CIFAR-$100$ \cite{krizhevsky2009learning}, Tiny ImageNet \cite{tinyimagenet} (i.e., the natural image datasets), and CRCHistoPhenotypes \cite{sirinukunwattana2016locality} (i.e., the medical image dataset).

\subsubsection{CIFAR-10}
The CIFAR-10 \cite{krizhevsky2009learning} is the most popular tiny image dataset consists of $10$ different categories of images, where each class has $6000$ images. The dimension of each image is $32\times32\times3$. To train the deep neural networks, we have used the training set (i.e., $50000$ images) of the CIFAR-10 dataset, and remaining data (i.e., $10000$ images) is utilized to validate the performance of the models. A few samples of images from the CIFAR-10 dataset are shown in Fig. \ref{image_samples}(a).

\subsubsection{CIFAR-100}
The CIFAR-100 \cite{krizhevsky2009learning} dataset is similar to CIFAR-10, except that CIFAR-100 has $100$ classes. In our experimental setting, the $50,000$ images are used to train the CNN models and the remaining $10,000$ images are used to validate the performance of the models. Similar to CIFAR-10, the dimension of each image is $32\times32\times3$. The sample images are shown in Fig. \ref{image_samples}(a).   

\subsubsection{Tiny ImageNet}
The Tiny ImageNet dataset \cite{tinyimagenet} consists a subset of ImageNet \cite{deng2009imagenet} images. This dataset has a total of $200$ classes and each class has $500$ training and $50$ validation images. In other words, we have used $100000$ images for training and $10000$ images for validating the performance of the models. The dimension of each image is $64\times64\times3$. The example images of the Tiny ImageNet dataset are portrayed in Fig. \ref{image_samples}(b).

\subsubsection{CRCHistoPhenotypes}
In order to generalize the observations reported in this paper, we have used a medical image dataset (consists of routine colon cancer nuclei cells) called ``CRCHistoPhenotypes'' \cite{sirinukunwattana2016locality}, which is publicly available\footnote{\url{https://warwick.ac.uk/fac/sci/dcs/research/tia/data/crchistolabelednucleihe}}. 
This colon cancer dataset consists a total of $22444$ nuclei patches that belong to the four different classes, namely, `Epithelial', `Inflammatory', `Fibroblast', and `Miscellaneous'. In total, $7722$ images belong to the `Epithelial' class, $5712$ images belong to the `Fibroblast' class, $6971$ images belong to the `Inflammatory' class, and the `Miscellaneous' class has remaining $2039$. The dimension of each nuclei patch is $32\times32\times3$.
For training the CNN models, $80\%$ of entire data ($i.e., 17955$ images) is utilized and remaining $20\%$ data (i.e., $4489$ images) is used to validate the performance of the models. The sample images are displayed in Fig. \ref{image_samples}(c).

\textbf{Deeper vs Wider datasets \cite{bansal2017s}}:
For any two datasets with roughly same number of images, one dataset is said to be deeper \cite{bansal2017s} than another dataset, if it has more number of images per class in the training set. The other dataset which has a lower number of images per class (i.e., more number of classes compared to another one) in the training set is called the wider dataset. For example, CIFAR-10 and CIFAR-100 \cite{krizhevsky2009learning}, both the datasets have $50000$ images in the training set. The CIFAR-10 is a deeper dataset since it has $5000$ images per class in the training set. On the other hand, the CIFAR-100 is wider dataset because it has only $500$ images per class.

\section{Results and Analysis}
\label{results_analysis}
We have conducted extensive experiments to observe the useful practices in deep learning for the usage of Convolutional Neural Networks (CNNs). The four CNN models discussed in section \ref{Developed_CNN_models} are implemented to perform the experiments on publicly available CIFAR-10/100, Tiny ImageNet, and CRCHistoPhenotypes datasets. The results in terms of the classification accuracy are reported in this paper.

\subsection{Impact of FC layers on the performance of the CNN w.r.t. to the depth of the CNN}
To observe the effect of deeper/shallow architectures on FC layers, initially, the CNN models are trained with a single FC (output) layer. Then another FC layer is added manually before the output (FC) layer to observe the gain/loss in the performance due to the addition of the new FC layer. The number of neurons is chosen (for newly added FC layer) starting from the number of classes to all multiples of $2$ (i.e, powers of $2$ such as $16$, $32$, etc.), which is greater than the number of classes and up to $4096$. For instance, in the case of CIFAR-10 dataset \cite{krizhevsky2009learning}, the experiments are conducted by varying the number of neurons in the newly added FC layer with $10, 16, 32, 64, ... , 4096$ number of neurons. In the next step, one more FC layer is added before the recently added FC layer. The number of neurons for newly added FC layer is chosen, ranging from the value for which best performance is obtained in the previous setting to $4096$. Suppose we obtained the best performance over CIFAR-10 using CNN-1 having two FC layers with $512$, $10$ neurons CIFAR-10. Then, we observed the performance of the model by adding another FC layer with $512, 1024, 2048,$ and $4096$ number of neurons. The details like the number of FC layers, number of neurons in each FC layer, best classification accuracies obtained for CIFAR-10 dataset using the four CNN models are shown in Table \ref{results_CIFAR10}. It is evident from Table \ref{results_CIFAR10} that the deeper architectures (i.e., CNN-3, CNN-2 with more convolution layers (require relatively less number of FC layers and also less number of neurons in FC layers compared to the shallow architecture (i.e., CNN-1 with $5$ $Conv$ layers) for CIFAR-10 dataset.

To generalize the above-mentioned observation, we have computed the results by varying the number of FC layers over other datasets and reported the best performance in Table \ref{results_findings}. From Table \ref{results_findings}, similar findings are noticed that the deeper architectures do not require more FC layers. On the other hand, the shallow architectures (such as CNN-1) require more FC layers in order to obtain better performance for any dataset. The reasoning for such a behavior is related to the type of features being learned by the $Conv$ layers. In general, CNN architecture learns the hierarchical features from raw images. Zeiler and Fergus \cite{zeiler2014visualizing} shown that the early layers learn the low-level features, whereas the deeper layers learn the high-level (problem specific) features. It means that the final $Conv$ layer of shallow architecture produces less abstract features as compared to the deeper architecture. Thus, the number of FC layers needed for shallow architecture is more as compared to the deeper architectures. To provide powerful evidence to the findings reported in this paper, we have conducted experiments by considering the half of the images (images belong to $100$ classes) of Tiny ImageNet dataset. We name this configuration as setting-1 (refer Table \ref{results_findings}).
We have also considered SVM (hinge) loss to compare the results that we obtained using the popular cross-entropy loss function. The CNN architectures through which the best validation is obtained (using FC layer structure reported in Table \ref{results_findings}) are trained using hinge loss. The same results are specified in the last row of Table \ref{results_findings}.



\begin{table*}[!t]
\begin{center}
\caption{The effect of depth of the CNN models (i.e., CNN-1, CNN-2, and CNN-3) on FC layers for the CIFAR-10 dataset is shown in this table. The best and $2^{nd}$ best accuracies are highlighted in bold and italic, respectively. For example, the CNN-2 model produces the best accuracy of $92.29\%$ for three FC layers with $4096$, $256$, and $10$ neurons and the $2^{nd}$ best accuracy of $92.02\%$ for two FC layers with $256$ and $10$ neurons.}
\label{results_CIFAR10}
\begin{adjustbox}{width=\columnwidth,center}
\begin{tabular}{| c | c | c | }
\hline
  \textbf{CNN-1} & \textbf{CNN-2} & \textbf{CNN-3}   \\
\hline

  Output FC layer (44.29)    & Output FC layer (91.46) & \textit{Output FC layer (92.05)}  ) 
    \\ \hline

  $10\times10$ $(88.67)$    & $10\times10$ (91.14)  & $10\times10$ (91.03)   \\ \hline

 $16\times10$ (88.72)    & $16\times10$ (91.58) & $16\times10$ (91.77)     \\ \hline

   $32\times10$ (88.93) & $\textit{32}\times\textit{10}$ \textit{(91.99)}  &  $32\times10$ (92.02)   \\ \hline

 $64\times10$ (89.72) & $64\times10$ (91.82) &  $64\times10$ (91.8)  \\ \hline

 $128\times10$ (89.2) & $128\times10$ (91.86) & $128\times10$ (89.2)  \\ \hline

 $256\times10$ (89.23) & \textit{$\textit{256}\times\textit{10}$ (92.02)} & $256\times10$ (89.23)   \\ \hline

 $512\times10$ (88.95) & $512\times10$ (90.98)  & $512\times10$ (91.78)   \\ \hline

 $1024\times10$ (89.56) & $1024\times10$ (91.54)  & $\textbf{1024}\times\textbf{10}$ \textbf{(92.22)}  \\ \hline

 $2048\times10$ (87.4) & $2048\times10$ (91.27)  & $2048\times10$ (91.59)  \\ \hline

 $4096\times10$ (86.27) & $4096\times10$ (87.51) & $4096\times10$ (90.68)   \\ \hline

 $64\times64\times10$ (89.35) & $256\times256\times10$  (91.97) & $1024\times1024\times10$ (91.27)   \\ \hline

 $128\times64\times10$ (89.71) & $512\times256\times10$ (91.92) & $2048\times1024\times10$ (91.43)   \\ \hline

 $256\times64\times10$ (89.79) & $1024\times256\times10$ (91.53) &  $4096\times1024\times10$ (91.94)  \\ \hline

 $512\times64\times10$ (89.88) & $2048\times256\times10$ (91.95) &  -  \\ \hline

 $1024\times64\times10$ (90) & $\textbf{4096}\times\textbf{256}\times\textbf{10}$ \textbf{(92.29)} &  -   \\ \hline

 $2048\times64\times10$ (90.28) & $4096\times4096\times256\times10$ (91.64) & -   \\ \hline

 $4096\times64\times10$ (90.59) &  - & -    \\ \hline
  \textbf{$\textbf{4096}\times\textbf{4096}\times\textbf{64}\times\textbf{10}$ (90.77)}  & - &  -   \\ \hline
 $\textit{4096}\times\textit{4096}\times\textit{4096}\times\textit{64}\times\textit{10}$ \textit{(90.74)} & - & -  \\ \hline
\end{tabular}
\end{adjustbox}
\end{center}
\end{table*}

\begin{table}[!t]
\begin{center}
\caption{The best validation accuracies obtained over CIFAR-10, CIFAR-100, Tiny ImageNet and CRCHistoPhenotypes datasets using four CNN models (i.e., CNN-1, CNN-2, and CNN-3) are depicted in this table. The results are presented in terms of the FC layer structures and validation accuracy.
}
\label{results_findings}
\resizebox{\columnwidth}{!}{%
\begin{tabular}{| c | c | c | c | p{5cm} |  c |  }
\hline

S.No. & Architecture & \multicolumn{4}{c|}{Dataset}  \\ \cline{3-6}
& & CIFAR-10 & CIFAR-100 & Tiny ImageNet & CRCHistoPhenoTypes \\ \hline

1 & CNN-1 & $4096\times4096\times64\times10$ $(90.77)$ & $4096\times4096\times2048\times100$ $(69.21)$ & $4096\times4096\times1024\times200$ $(50.1)$

& $2048\times256\times4$ $(82.53)$ \\ \hline

2 & CNN-2 & $4096\times256\times10$  $(92.29)$ &	$4096\times100$  $(62.28)$ &	$512\times200$, $1024\times200$ $(41.84)$ 


&	$512\times4$  $(84.89)$ \\ \hline

3 & CNN-3 & $1024\times10$ $(92.22)$
&	Single FC (output) layer $(66.98)$ &	Single FC (output) layer $(40.27)$	

& $128x4$  $(84.94)$ \\ \hline


\end{tabular}%
}
 
 \end{center}
\end{table}

\subsection{Effect of FC layers on the performance of the CNN model w.r.t. to different types of datasets}
We have used two kinds of datasets (deeper and wider) to analyze the effect of FC layers on the performance of CNN. 
Table \ref{factors_table} presents the characteristics like average number of images per class in the training set ($N$), number of classes ($C$), number of training images ($Tr$), and validation images ($Va$) of four datasets discussed in section \ref{datasets}.

\begin{figure}[!t]
\centering
\includegraphics[width=1 \textwidth]{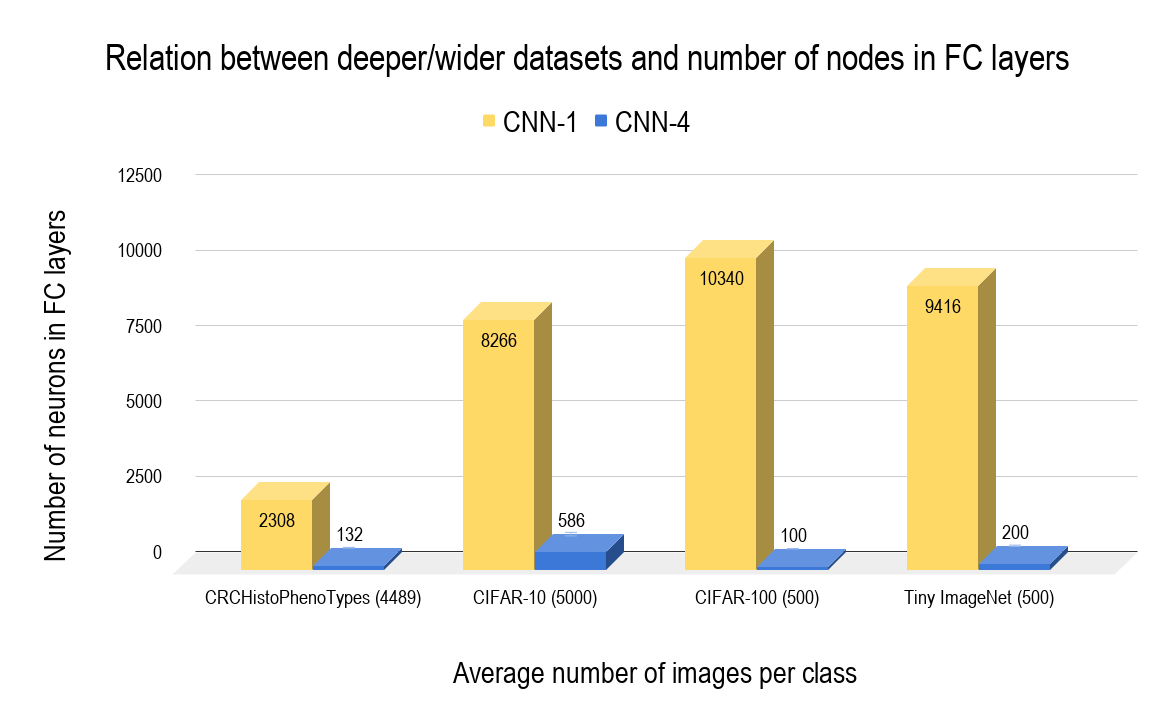}
\caption{The effect of deeper/wider datasets on FC layers of CNN. For wider datasets, deeper architecture (CNN-3) requires relatively less number of neurons in FC layers than deeper datasets. On the other hand, for wider datasets, shallow architecture (CNN-1) requires relatively large number of neurons in FC layers compared to deeper datasets.}
\label{impact_FClayers_datasets}
\end{figure}

From Fig. \ref{impact_FClayers_datasets}, we can observe that shallow architecture CNN-1 (less deeper than CNN-2, and CNN-3) requires more neurons in FC layers for wider datasets compared to deeper datasets. On the other hand, deeper architecture CNN-3 (deeper than CNN-1) requires fewer neurons in FC layers for wider datasets compared to deeper datasets. Deeper CNN models such as CNN-3, CNN-2 have more number of trainable parameters in $Conv$ layers. Thus, a deeper dataset is required to learn large parameters of the network. In contrast, a shallow architecture like CNN-1 with $5$ $Conv$ layers has fewer parameters for which a wider dataset is better suited to train the model.

\begin{table}[!t]
\begin{center}
\caption{The characteristics of CIFAR-10, CIFAR-100, Tiny ImageNet, and CRCHistoPhenotypes datasets are presented in this table. Here, $N$ represents the average number of images per class in the training set, $C$ represents the number of classes corresponding to a dataset, $Tr$ and $Va$ are the number of images in the Training and Validation sets, respectively.}
\label{factors_table}
\begin{tabular}{| c | c | c | c | c |}
\hline
\textbf{Dataset}  
 & \textbf{N} & \textbf{C} & \textbf{Tr} & \textbf{Va}\\
\hline
CIFAR-10  &    5000  & 10 & 50,000 & 10,000\\ \hline
CIFAR-100 &    500 & 100 & 50,000 & 10,000\\ \hline
Tiny ImageNet   & 500 & 200 & 80,000 & 20,000\\ \hline
CRCHistoPhenotypes  &  4489    & 4 & 17955 & 4489\\ \hline
\end{tabular}
\end{center}
\end{table}

\subsection{Deeper vs. Shallower Architectures, Which are better and when?}
\label{deep/shallow}
Bansal \etal \cite{bansal2017s} have reported that the deeper architectures are preferred over shallow architectures while training the CNN models with deeper datasets. Whereas, for the wider datasets, the shallow architectures perform better compared to the deeper architectures. However, this observation is specific to face recognition problem as reported in \cite{bansal2017s}. In this paper, we made a rigorous study to generalize this finding by conducting extensive experiments on different modalities of datasets. For example, CIFAR-10, CIFAR-100, and Tiny ImageNet datasets have the natural images and the CRCHistoPhenotypes dataset has the medical images. The results obtained through these experiments clearly indicate that the deeper architectures are always preferred over shallow architectures to train the CNN model using deeper datasets. In contrast, for the wider datasets, the shallow architectures perform better than the deeper CNN models. 

From Table \ref{results_findings}, we can observe that training deeper architectures CNN-2 and CNN-3 with deeper dataset produce $92.29\%$ and $92.22\%$ validation accuracies for the CIFAR-10 dataset and $84.89\%$ and $84.94\%$ for the CRCHistoPhenotypes dataset. In contrast, we obtained $90.77\%$ and $82.53\%$ validation accuracies, when the shallow architecture CNN-1 is trained with CIFAR-10 and CRCHistoPhenotypes datasets, respectively. On the other hand, for the wider datasets such as CIFAR-100 and Tiny ImageNet, better performance is obtained using the shallow architecture (CNN-1). From Table \ref{results_findings}, we can observe that the CNN-1 gives a validation accuracy of $69.21\%$ for CIFAR-100 and $50.1\%$ for Tiny ImageNet dataset. Whereas, the CNN-1 model performs relatively poor for deeper datasets. 

This observation is very much useful while choosing a CNN architecture to train the model for a given dataset. The generalization of this finding intuitively makes sense because the deeper/shallow architectures have a more/less number of trainable parameters, in a typical CNN model which require more/less number of images per subject (class) for the training.

\section{Conclusion}
\label{conclusion}
In this paper, we have analyzed the effect of certain decisions in terms of the FC layers of CNN for image classification. Careful selection of these decisions not only improves the performance of the CNN models but also reduces the time required to choose among different architectures such as deeper and shallow. This paper is concluding the following guidelines that can be adopted while designing the deep/shallow convolutional neural networks to obtain better performance.
\begin{itemize}
\item In order to obtain better performance, the shallow CNNs require more nodes in FC layers. On the other hand, deeper CNNs need less number of neurons in FC layers irrespective of type of the dataset.

\item The shallow CNNs require a large number of neurons in FC layers as well as more number of FC layers for \textbf{wider datasets} compared to \textbf{deeper datasets} and vice-versa.

\item Deeper CNNs perform better than shallow models over \textbf{deeper datasets}. In contrast, shallow architectures perform better than deeper architectures for \textbf{wider datasets}. These observations can help the deep learning community while making a decision about the choice of deep/shallow CNN architectures.
\end{itemize}

\section*{ACKNOWLEDGMENT}
This work is supported in part by the Science and Engineering Research Board (SERB), Govt. of India, Grant No. ECR/2017/000082. We gratefully acknowledge the support of NVIDIA Corporation with the donation of the GeForce Titan XP GPU. 

\section*{References}

\bibliography{mybibfile}

\end{document}